\pdfoutput=1
\documentclass[10pt, conference, compsocconf]{IEEEtran}

\usepackage{common}
\usepackage{nohyperref}
\hypersetup{bookmarks=false}
\usepackage{url}  

\graphicspath{{figures/}}
\newcommand{\eg}{\emph{e.g.}}
\newcommand{\ie}{\emph{i.e.}}
\newcommand{\etal}{\emph{et~al.}}

\begin{document}
\title{Distributed Deep Neural Networks\\over the Cloud, the Edge and End Devices}

\author{
 \IEEEauthorblockN{Surat Teerapittayanon}
 \IEEEauthorblockA{Harvard University\\
 Cambridge, MA, USA\\
 Email: steerapi@seas.harvard.edu}
 \and
 \IEEEauthorblockN{Bradley McDanel}
 \IEEEauthorblockA{Harvard University\\
 Cambridge, MA, USA\\
 Email: mcdanel@fas.harvard.edu}
 \and
 \IEEEauthorblockN{H.T. Kung}
 \IEEEauthorblockA{Harvard University\\
 Cambridge, MA, USA\\
 Email: kung@harvard.edu}
}

\maketitle

\begin{abstract}
We propose distributed deep neural networks (DDNNs) over distributed computing hierarchies, consisting of the cloud, the edge (fog) and end devices. While being able to accommodate inference of a deep neural network (DNN) in the cloud, a DDNN also allows fast and localized inference using shallow portions of the neural network at the edge and end devices. When supported by a scalable distributed computing hierarchy, a DDNN can scale up in neural network size and scale out in geographical span. Due to its distributed nature, DDNNs enhance sensor fusion, system fault tolerance and data privacy for DNN applications.  In implementing a DDNN, we map sections of a DNN onto a distributed computing hierarchy. By jointly training these sections, we minimize communication and resource usage for devices and maximize usefulness of extracted features which are utilized in the cloud. The resulting system has built-in support for automatic sensor fusion and fault tolerance. As a proof of concept, we show a DDNN can exploit geographical diversity of sensors to improve object recognition accuracy and reduce communication cost. In our experiment, compared with the traditional method of offloading raw sensor data to be processed in the cloud, DDNN locally processes most sensor data on end devices while achieving high accuracy and is able to reduce the communication cost by a factor of over 20x.
\end{abstract}


\IEEEpeerreviewmaketitle

\section{Introduction}
\begin{figure}
    \centering
    \includegraphics[width=\linewidth]{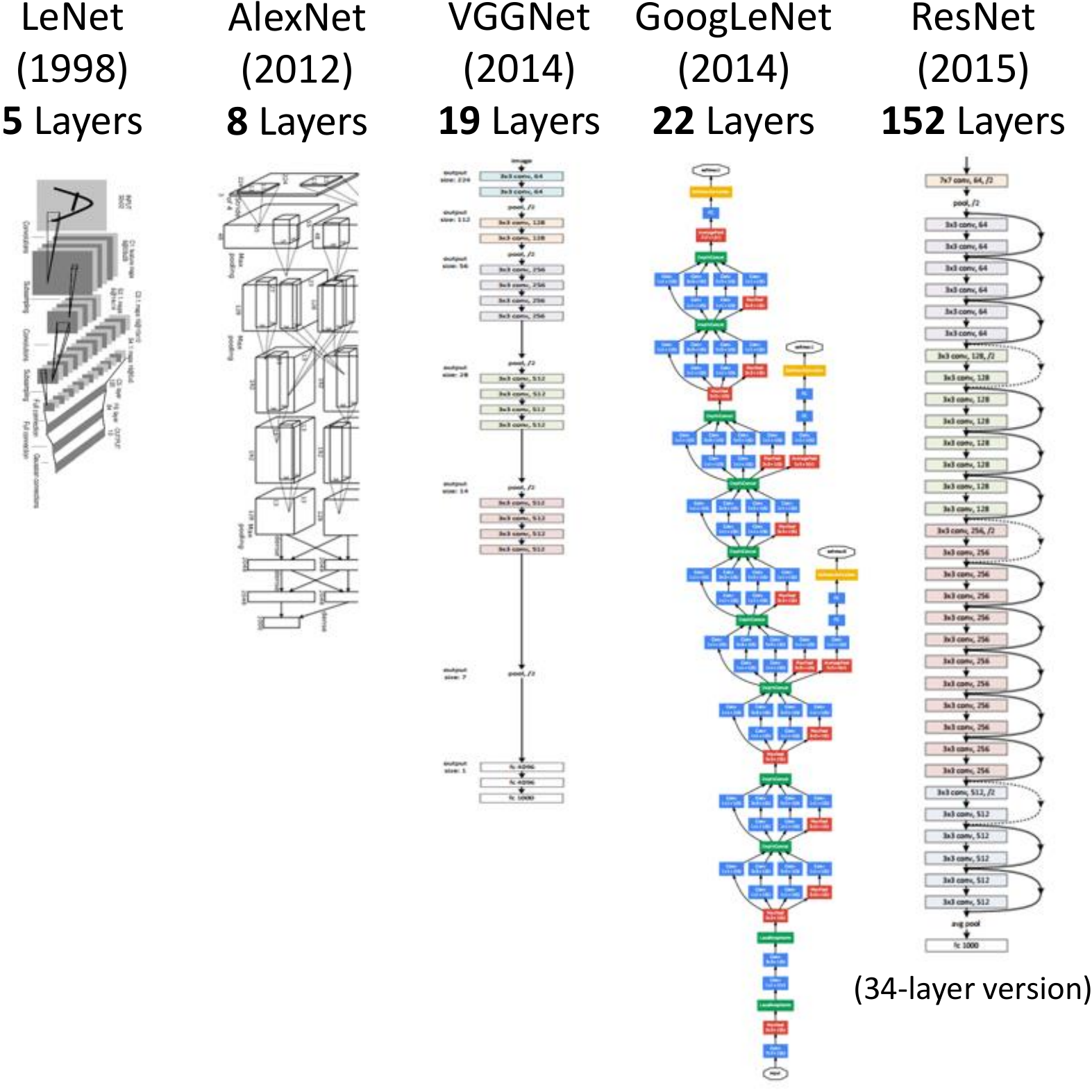}
    \caption{Progression towards deeper neural network structures in recent years (see,~\eg,~\cite{lecun1998gradient,krizhevsky2012imagenet,simonyan2014very,szegedy2015going,he2015deep}).}
    \label{fig:deeper}
\end{figure}

Neural networks (NNs), and deep neural networks (DNNs) in particular, have achieved great success in numerous applications in recent years. For example, deep Convolutional Neural Networks (CNNs) continuously achieve state-of-the-art performances on various tasks in computer vision as shown in Figure~\ref{fig:deeper}. At the same time, the number of end devices, including Internet of Things (IoT) devices, has increased dramatically. These devices are appealing targets for machine learning applications as they are often directly connected to sensors (\eg,~cameras, microphones, gyroscopes) that capture a large quantity of input data in a streaming fashion.

However, the current state of machine learning systems on end devices leaves an unsatisfactory choice: either (1) offload input sensor data to large NN models (\eg,~DNNs) in the cloud, with the associated communication costs, latency issues and privacy concerns, or (2) perform classification directly on the end device using simple Machine Learning (ML) models \eg,~linear Support Vector Machine (SVM), leading to reduced system accuracy.

To address these shortcomings, it is natural to consider the use of a distributed computing approach. Hierarchically distributed computing structures consisting of the cloud, the edge and devices~(see, \eg,~\cite{shiedge, skala2015scalable})~have inherent advantages, such as supporting coordinated central and local decisions, and providing system scalability, for large-scale intelligent tasks based on geographically distributed IoT devices. 

An example of one such distributed approach is to combine a small NN\footnote{The term network layer may refer to either a layer in a NN or a layer in the distributed computing hierarchy (\eg,~edge or cloud). In order to remove ambiguity, when we refer to network layers for NN we explicitly use the term NN layers.} model (less number of parameters) on end devices and a larger NN model (more number of parameters) in the cloud. The small model at an end device can quickly perform initial feature extraction, and also classification if the model is confident. Otherwise, the end device can fall back to the large NN model in the cloud, which performs further processing and final classification. This approach has the benefit of low communication costs compared to always offloading NN input to the cloud and can achieve higher accuracy compared to a simple model on device. Additionally, since a summary based on extracted features from the end device model are sent instead of raw sensor data, the system could provide better privacy protection.

However, this kind of distributed approach over a computing hierarchy is challenging for a number of reasons, including:


\begin{itemize}

    \item End devices such as embedded sensor nodes often have limited memory and battery budgets. This makes it an issue to fit models on the devices that meet the required accuracy and energy constraints.
    
    \item A straightforward partitioning of NN models over a computing hierarchy may incur prohibitively large communication costs in transferring intermediate results between computation nodes.
    
    \item Incorporating geographically distributed end devices is generally beyond the scope of DNN literature. When multiple sensor inputs on different end devices are used, they need to be aggregated together for a single classification objective. A trained NN will need to support such sensor fusion.
    
    \item Multiple models at the cloud, the edge and the device need to be learned jointly to allow coordinated decision making. Computation already performed on end device models should be useful for further processing on edge or cloud models.
        
    \item Usual layer-by-layer processing of a DNN from the NN's input layer all the way to the NN's output layer does not directly provide a mechanism for local and fast inference at earlier points in the neural networks (e.g., end devices).
        
    \item A balance is needed between the accuracy of a model (with the associated model size) at a given distributed computing layer and the cost of communicating to the layer above it. The solution must have reasonably good lower NN layers on the end devices capable of accurate local classification for some input while also providing useful features for classification in the cloud for other input.

    
\end{itemize}

To address these concerns under the same optimization framework, it is desirable that a system could train a \textit{single} end-to-end model, such as a DNN, and partition it between end devices and the cloud\footnote{For presentation simplicity, we often just consider the device-cloud scenario. Our methodology can similarly apply to general device-edge (fog)-cloud scenarios.}, in order to provide a simpler and more principled approach.

To this end, we propose distributed deep neural networks (DDNNs) over distributed computing hierarchies, consisting of the cloud, the edge (fog) and geographically distributed end devices. In implementing a DDNN, we map sections of a single DNN onto a distributed computing hierarchy. By jointly training these sections, we show that DDNNs can effectively address the aforementioned challenges. Specifically, while being able to accommodate inference of a DNN in the cloud, a DDNN allows fast and localized inference using some shallow portions of the DNN at the edge and end devices. Moreover, via distributed computing, DDNNs naturally enhance sensor fusion, data privacy and system fault tolerance for DNN applications. When supported by a scalable distributed computing hierarchy, a DDNN can scale up in neural network size and scale out in geographical span. 

DDNN leverages our earlier work on BranchyNet~\cite{teerapittayanon2016branchynet} which allows early exit points to be placed in a DNN. Samples can be classified and exited locally when the system is confident and offloaded to the edge and the cloud when additional processing is required. In addition, DDNN leverages the recent work of binary neural networks (BNNs)~\cite{courbariaux2015binaryconnect}, which greatly reduce the required memory cost of neural network layers and enables multi-layer NNs to run on end devices with small memory footprints~\cite{mcdanel2016ebnn}.   By training DDNN end-to-end, the network optimally configures lower NN layers to support local inference at end devices, and higher NN layers in the cloud to improve overall classification accuracy of the system. As a proof of concept, we show a DDNN can exploit geographical diversity of sensors (on a multi-view multi-camera dataset) in sensor fusion to improve recognition accuracy. 

The contributions of this paper include 
\begin{enumerate}
    \item A novel DDNN framework and its implementation that maps sections of a DNN onto a distributed computing hierarchy.
    \item A joint training method that minimizes communication and resource usage for devices and maximizes usefulness of extracted features which are utilized in the cloud, while allowing low-latency classification via early exit for a high percentage of input samples.
    \item Aggregation schemes that allows automatic sensor fusion of multiple sensor inputs to improve the overall performance (accuracy and fault tolerance) of the system.
\end{enumerate} 

The DDNN codebase is open source and can be found here: \url{https://github.com/kunglab/ddnn}.


\section{Related Work}
In this section, we briefly review related work in distributed computing hierarchies and recent deep learning algorithms that enable our proposed method to run in a distributed fashion. We then discuss other approaches involving distributed deep networks.  

\subsection{Distributed Computing Hierarchy}
The framework of a large-scale distributed computing hierarchy has assumed new significance in the emerging era of IoT. It is widely expected that most of data generated by the massive number of IoT devices must be processed locally at the devices or at the edge, for otherwise the total amount of sensor data for a centralized cloud would overwhelm the communication network bandwidth. In addition, a distributed computing hierarchy offers opportunities for system scalability, data security and privacy, as well as shorter response times (see,~\eg,~\cite{skala2015scalable, yi2015fog}). For example, in~\cite{yi2015fog}, a face recognition application shows a reduced response time is achieved when a smartphone's photos are proceeded by the edge (fog) as opposed to the cloud. In this paper, we show that DDNN can systematically exploit the inherent advantages of a distributed computing hierarchy for DNN applications and achieve similar benefits.

\subsection{Deep Neural Network Extensions}
Binarized neural networks (BNNs) are a recent type of neural networks, where the weights in linear and convolutional layers are constrained to $\{-1, 1\}$ (stored as $0$ and $1$ respectively). This representation has been shown to achieve similar classification accuracy for some datasets such as MNIST and CIFAR-10~\cite{rastegari2016xnor} when compared to a standard floating-point neural network while using less memory and reduced computation due to the binary format~\cite{courbariaux2015binaryconnect}. Embedded binarized neural networks (eBNNs) extends BNNs to allow the network to fit on embedded devices by reducing floating-point temporaries through reordering the operations in inference~\cite{mcdanel2016ebnn}. These compact models are especially attractive in end device settings, where memory can be a limiting factor and low power consumption is required. In DDNN, we use BNNs, eBNNs and the alike to accommodate the end devices, so that they can be jointly trained with the NN layers in the edge and cloud.

BranchyNet proposed a solution of classifying samples at earlier points in a neural network, called early exit points, through the use of an entropy-based confidence criteria~\cite{teerapittayanon2016branchynet}. If at an early exit point a sample is deemed confident based on the entropy of the computed probability vector for target classes, then it is classified and no further computation is performed by the higher NN layers. In DDNN, exit points are placed at physical boundaries (\eg,~between the last NN layer on an end device and the first NN layer in the next higher layer of the distributed computing hierarchy such as the edge or the cloud). Input samples that can already be classified early will exit locally, thereby achieving a lowered response latency and saving communication to the next physical boundary. With similar objectives, SACT~\cite{figurnov2016sact} allocates computation on a per region basis in an image, and exits each region independently when it is deemed to be of sufficient quality. 

\subsection{Distributed Training of Deep Networks}
Current research on distributing deep networks is mainly focused on improving the runtime of training the neural network. In 2012, Dean~\etal~proposed DistBelief, which maps large DNNs over thousands of CPU cores during training~\cite{dean2012large}. More recently, several methods have been proposed to scale up DNN training across GPU clusters~\cite{iandola2015firecaffe, dean2015large}, which further reduces the runtime of network training. Note that this form of distributing DNNs (over homogeneous computing units) is fundamentally different from the notion presented in this paper. We proposes a way to train and perform feedforward inference over deep networks that can be deployed over a distributed computing hierarchy, rather than processed in parallel over bus- or switch-connected CPUs or GPUs in the cloud.

\section{Proposed Distributed Deep Neural Networks}
In this section we give an overview of the proposed distributed deep neural network (DDNN) architecture and describe how training and inference in DDNN is performed.

\subsection{DDNN Architecture}
\begin{figure*}
    \centering
    \includegraphics[width=\linewidth]{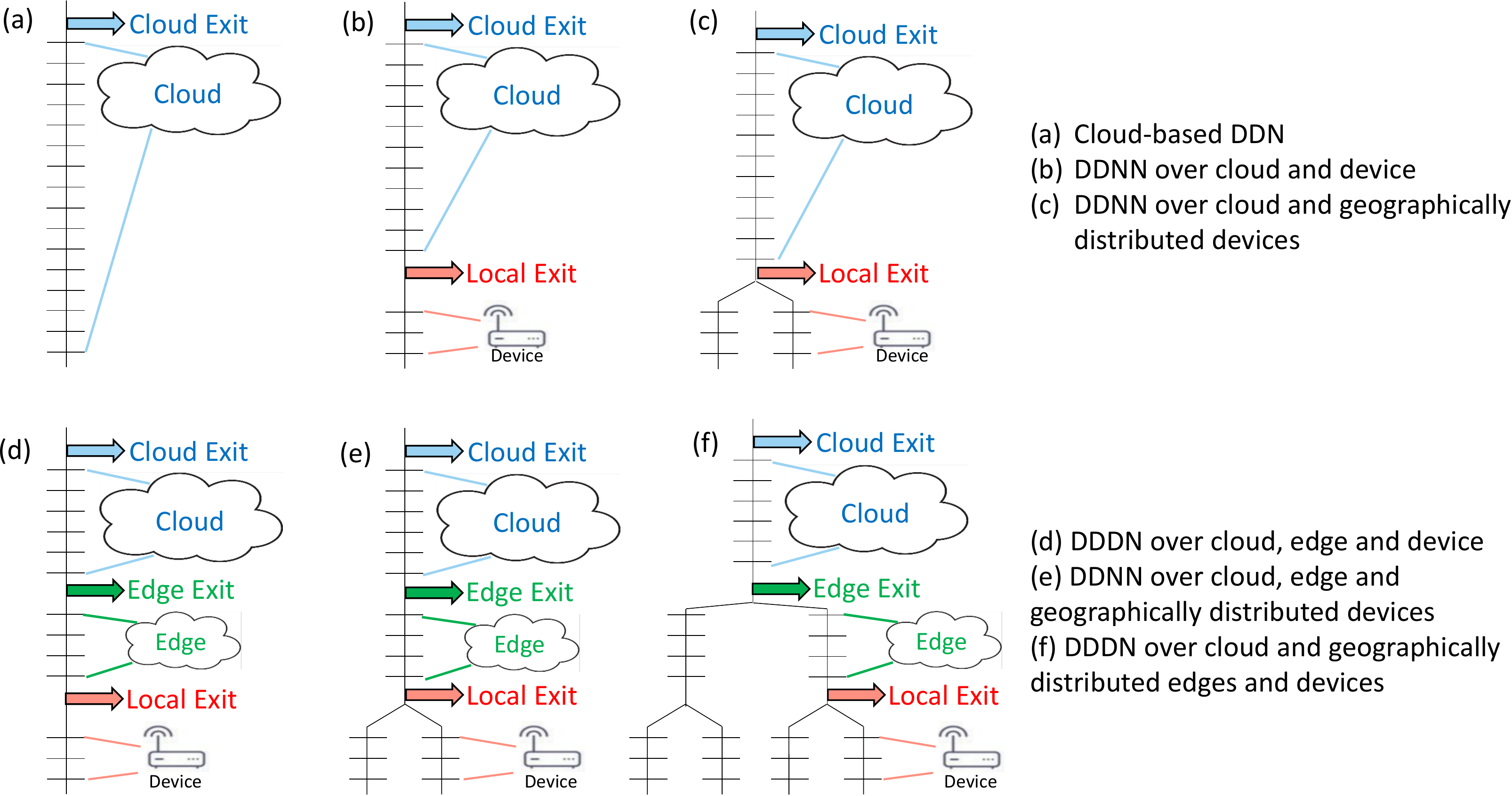}
    \caption{Overview of the DDNN architecture. The vertical lines represent the DNN pipeline, which connects the horizontal bars (NN layers). (a) is the standard DNN (processed entirely in the cloud), (b) introduces end devices and a local exit point that may classify samples before the cloud, (c) extends (b) by adding multiple end devices which are aggregated together for classification, (d) and (e) extend (b) and (c) by adding edge layers between the cloud and end devices, and (f) shows how the edge can also be distributed like the end devices.}
    \label{fig:ddnn}
\end{figure*}

DDNN maps a trained DNN onto heterogeneous physical devices distributed locally, at the edge, and in the cloud. Since DDNN relies on a jointly trained DNN framework at all parts in the neural network, for both training and inference, many of the difficult engineering decisions are greatly simplified. Figure~\ref{fig:ddnn}~provides an overview of the DDNN architecture. The configurations presented show how DDNN can scale the inference computation across different physical devices. The cloud-based DDNN in (a) can be viewed as the standard DNN running in the cloud as described in the introduction. In this case, sensor input captured on end devices is sent to the cloud in original format (raw input format), where all layers of DNN inference is performed. 

We can extend this model to include a single end device, as shown in (b), by performing a portion of the DNN inference computation on the device rather than sending the raw input to the cloud. Using an exit point after device inference, we may classify those samples which the local network is confident about, without sending any information to the cloud. For more difficult cases, the intermediate DNN output (up to the local exit) is sent to the cloud, where further inference is performed using additional NN layers and a final classification decision is made. Note that the intermediate output can be designed to be much smaller than the sensor input (\eg,~a raw image from a video camera), and therefore drastically reduce the network communication required between the end device and the cloud. The details of how communication is considered in the network is discussed in section~\ref{sec:ddnn-comm}.

DDNN can also be extended to multiple end devices which may be geographically distributed, shown in (c), that work together to make a classification decision. Here, each end device performs local computation as in (b), but their output is aggregated together before the local exit point. Since the entire DDNN is jointly trained across all end devices and exit points, the network automatically aggregates the input with the objective of achieving maximum classification accuracy. This automatic data fusion (sensor fusion) simplifies runtime inference by avoiding the necessity of manually combining output from multiple end devices. We will discuss the design of feature aggregation in detail in section~\ref{sec:agg}. As before, if the local exit point is not confident about the sample, each end devices sends intermediate output to the cloud, where another round of feature aggregation is performed before making a final classification decision.

DDNN scales vertically as well, by using an edge layer in the distributed computing hierarchy between the end devices and cloud, shown in (d) and (e). The edge acts similarly to the cloud, by taking output from the end devices, performing aggregation and classification if possible, and forwarding its own intermediate output to the cloud if more processing is needed. In this way, DDNN naturally adjusts the network communication and response time of the system on a per sample basis. Samples that can be correctly classified locally are exiting without any communication to the edge or cloud. Samples that require more feature extraction than can be provided locally are sent to the edge, and eventually the cloud if necessary. Finally, DDNNs can also scale geographically across the edge layer as well, which is shown in (f).


\subsection{DDNN Aggregation Methods}
\label{sec:agg}
In DDNN configurations with multiple end devices (\eg,~(c), (e), and (f) in Figure~\ref{fig:ddnn}), the output from each end device must be aggregated in order to perform classification. We present several different schemes for aggregating the output. Each aggregation method makes different assumptions about how the device output should be combined and therefore can result in different system accuracy. We present three approaches:
\begin{itemize}
    \item Max pooling (MP). MP aggregates the input vectors by taking the max of each component. Mathematically, max pooling can be written as 
    $$ {\hat v}_{j} = \max_{1 \leq i \leq n} v_{ij}, $$
    where $n$ is the number of inputs and $v_{ij}$ is the $j$-th component of the input vector and ${\hat v}_{j}$ is the $j$-th component of the resulting output vector.
    \item Average pooling (AP). AP aggregates the input vectors by taking the average of each component. This is written as
    $$ {\hat v}_{j} = \sum_{i=1}^n \frac{v_{ij}}{n}, $$
    where $n$ is the number of inputs and $v_{ij}$ is the $j$-th component of the input vector and ${\hat v}_{j}$ is the $j$-th component of the resulting output vector. Averaging may reduce noisy input presented in some end devices.  
    \item Concatenation (CC). CC simply concatenates the input vectors together. CC retains all information which is useful for higher layers (\eg,~the cloud) that can use the full information to extract higher level features. Note that this expands the dimension of the resulting vector. To map this vector back to the same number of dimensions as input vectors, we add an additional linear layer.
\end{itemize}

We analyzes these aggregation methods in Section~\ref{sec:agganalysis}.

\subsection{DDNN Training}
\label{sec:training}
While DDNN inference is distributed over the distributed computing hierarchy, the DDNN system can be trained on a single powerful server or in the cloud. One aspect of DDNN that is different from most conventional DNN pipelines is the use of multiple exit points as shown in Figure~\ref{fig:ddnn}. At training time, the loss from each exit is combined during back-propagation so that the entire network can be jointly trained, and each exit point achieves good accuracy relative to its depth. For this work, we follow joint training as described in GoogleNet~\cite{szegedy2015going} and BranchyNet~\cite{teerapittayanon2016branchynet}.

For the system evaluation discussed in Section~\ref{sec:eval}, we apply DDNNs to a classification task. We use the softmax cross entropy loss function as the optimization objective. We now describe formally how we train DDNNs. Let $\boldy$ be a one-hot ground-truth label vector, $\boldx$ be an input sample and $\mathcal C$ be the set of all possible labels. For each exit, the softmax cross entropy objective function can be written as
\begin{align*}
L(\hat{\boldy}, \boldy;\theta) = 
&-\frac{1}{|\mathcal C|} \displaystyle \sum_{c\in \mathcal C} y_c \log \hat{y_c},
\end{align*}
where 
\begin{align*}
\hat{\boldy}
&= \text{softmax}(\boldz) 
= \frac{\exp(\boldz)}{\displaystyle \sum_{c\in \mathcal C} \exp(z_c)},
\end{align*}
and
\begin{align*}
\boldz =
& f_{\text{exit}_n}(\boldx;\theta),
\end{align*}
where $f_{\text{exit}_n}$ is a function representing the computation of the neural network layers from an entry point to the $n$-th exit branch and $\theta$ represents the network parameters such as weights and biases of those layers.

To train the DDNN we form a joint optimization problem as minimizing a weighted sum of the loss functions of each exit:
\begin{align*}
    L(\hat{\boldy}, \boldy;\theta) = \displaystyle \sum_{n=1}^N w_n L(\hat{\boldy}_{\text{exit}_n}, \boldy;\theta),
\end{align*}
where $N$ is the total number of exit points and $w_n$ is the associated weight of each exit. Equal weights are used for the experimental results of this paper. 

\subsection{DDNN Inference}

Inference in DDNN is performed in several stages using multiple preconfigured exit thresholds $\boldT$ (one element $T$ at each exit point) as a measure of confidence in the prediction of the sample. One way to define $\boldT$ is by searching over the ranges of $\boldT$ on a validation set and pick the one with the best accuracy. We use a normalized entropy threshold as the confidence criteria (instead of unnormalized entropy as used in~\cite{teerapittayanon2016branchynet}) that determines whether to classify (exit) a sample at a particular exit point. The normalized entropy is defined as

$$\eta (\mathbf{x}) =-\sum_{i=1}^{|\mathcal{C}|}{\frac {x_{i}\log x_{i}}{\log |\mathcal{C}|}},$$
where $\mathcal C$ is the set of all possible labels and $\textbf{x}$ is a probability vector. This normalized entropy $\eta$ has values between 0 and 1 which allows easier interpretation and searching of its corresponding threshold $T$. For example, $\eta$ close to 0 means that the DDNN is confident about the prediction of the sample; $\eta$ close to 1 means it is not confident. At each exit point, $\eta$ is computed and compared against $T$ in order to determine if the sample should exit at that point.

At a given exit point, if the predictor is not confident in the result (\ie,~$\eta > T$), the system falls back to a higher exit point in the hierarchy until the last exit is reached which always performs classification.

We now provide an example of the inference procedure for a DDNN which has multiple end devices and three exit points (configuration (e) in Figure~\ref{fig:ddnn}):
\begin{enumerate}
  \item Each end device first sends summary information to local aggregator.
  \item The local aggregator determines if the combined summary information is sufficient for accurate classification.
  \item If so, the sample is classified (exited). 
  \item If not, each device sends more detailed information to the edge in order to perform further processing for classification.
  \item If the edge believes it can correctly classify the sample it does so and no information is sent to the cloud.
  \item Otherwise, the edge forwards intermediate computation to the cloud which makes the final classification.
\end{enumerate}

\subsection{Communication Cost of DDNN Inference}
\label{sec:ddnn-comm}
The total communication cost for an end device with the local and cloud aggregator is calculated as 
\begin{equation}
\label{eq:comm}
c = 4\times |\mathcal{C}| + (1-l)\frac{f\times o}{8}    
\end{equation}
where $l$ is the percentage of samples exited locally, $\mathcal{C}$ is the set of all possible labels (3 in our experiments), $f$ is the number of filters, and $o$ is the output size of a single filter for the final NN layer on the end-device. The constant 4 corresponds to 4 bytes which are used to represent a floating-point number and the constant 8 corresponds to bits used to express a byte output. The first term assumes a single floating-point per class, which conveys the probability that the sample to be transmitted from the end device to the local aggregator belongs to this class. This step happens regardless of whether the sample is exited locally or at a later exit point. The second term is the communication between end device and cloud which happens $(1-l)$ fraction of the time, when the sample is exited in the cloud rather than locally.

\subsection{Accuracy Measures}
\label{sec:acc-def}
Throughout the evaluation in Section~\ref{sec:eval}, we use different accuracy measures for the various exit points in a DDNN as follows:
\begin{itemize}
    \item \textit{Local Accuracy} is the accuracy when exiting 100\% of samples at the local exit of a DDNN.
    \item \textit{Edge Accuracy} is the accuracy when exiting 100\% of samples at the edge exit of a DDNN.
    \item \textit{Cloud Accuracy} is the accuracy when exiting 100\% of samples at the cloud exit of a DDNN.
    \item \textit{Overall Accuracy} is the accuracy when exiting some percentage of samples at each exit point in the hierarchy. The samples classified at each exit point are determined by the entropy threshold $T$ for that exit. The impact of $T$ on classification accuracy and communication cost is discussed in Section~\ref{sec:entropy}.
    \item \textit{Individual Accuracy} is the accuracy of an end device NN model trained separately from DDNN. The NN model for each end device consists of a ConvP block followed by a FC block (a single end device portion as shown in Figure~\ref{fig:merge}). In the evaluation, individual accuracy for each device is computed by classifying all samples using the individual NN model and not relying on the local or cloud exit points of a DDNN.
\end{itemize}

\section{DDNN System Evaluation}
\label{sec:eval}
In this section, we evaluate DDNN on a scenario with multiple end devices and demonstrate the following characteristics of the approach:
\begin{itemize}
    \item DDNNs allow multiple end devices to work collaboratively in order to improve accuracy at both the local and cloud exit points.
    \item DDNNs seamlessly extend the capability of end devices by offloading difficult samples to the cloud.
    \item DDNNs have built-in fault tolerance. We illustrate that missing any single end device does not dramatically affect the accuracy of the system. Additionally, we show how performance gradually degrades as more end devices are lost.
    \item DDNNs reduce communication costs for end devices compared to traditional system that offloads all input sensor data to the cloud.
\end{itemize}

We first introduce the DDNN architecture and dataset used in our evaluation.

\subsection{DDNN Evaluation Architecture}
\begin{figure}
    \centering
    \includegraphics[width=0.7\linewidth]{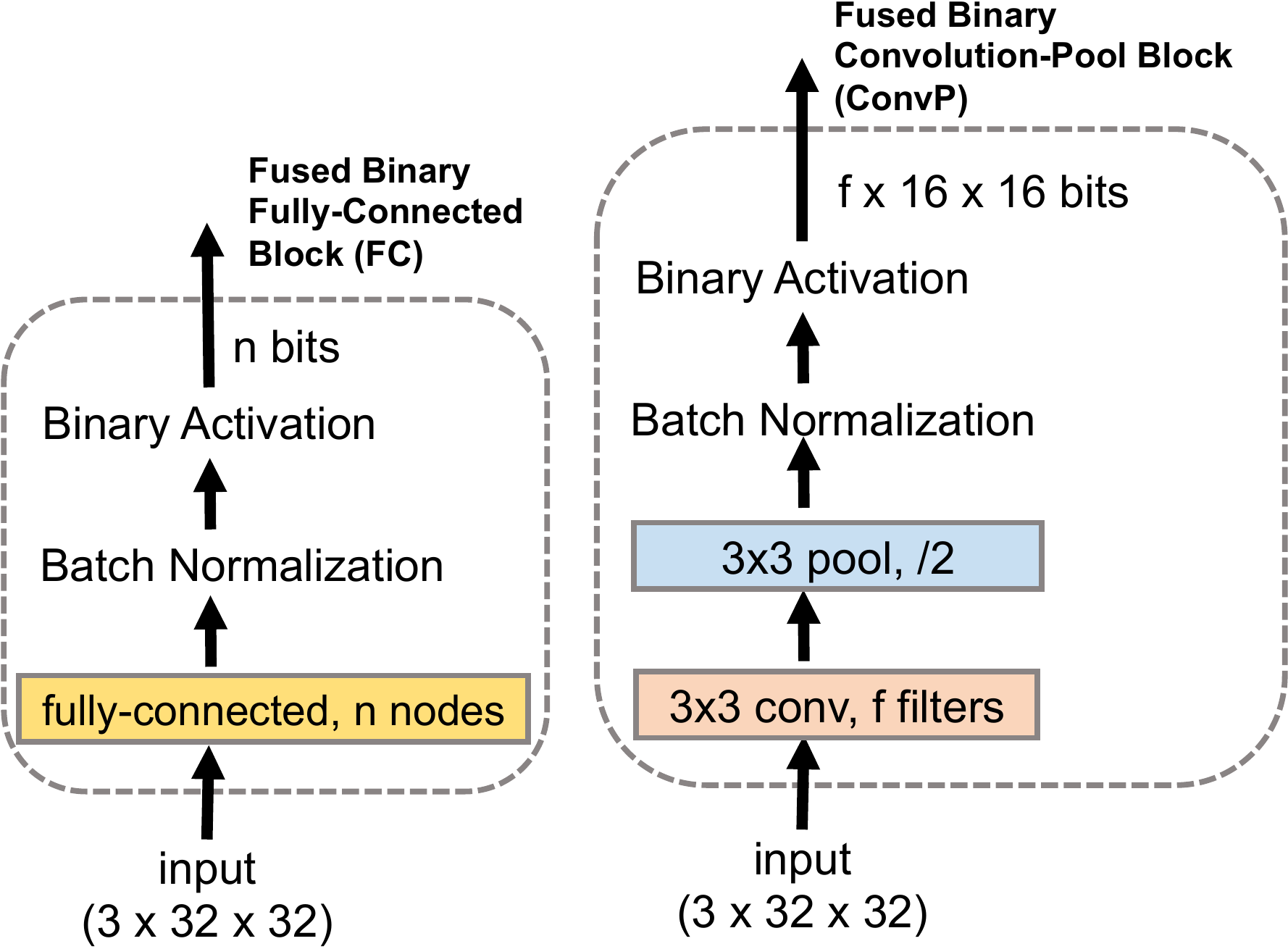}
    \caption{Fused binary blocks consisting of one or more standard NN layers. The fused binary fully connected (FC) block is a fully connected layer with $n$ nodes, batch normalization and binary activation. The fused binary convolution-pool (ConvP) block consists of a convolutional layer with $f$ filters, a pooling layer, batch normalization and binary activation. The convolution layer has a kernel of size 3x3 with stride 1 and padding 1. The pooling layer has a kernel of size 3x3 with stride 2 and padding 1. These blocks are used as they are presented in~\cite{mcdanel2016ebnn}.}
    \label{fig:fused}
\end{figure}

To accommodate the small memory size of the end devices, we use Binary Neural Network~\cite{courbariaux2015binaryconnect} blocks\footnote{A block consists of one or more conventional NN layers}. We make use of two types of blocks in~\cite{mcdanel2016ebnn}: the fused binary fully connected (FC) block and fused binary convolution-pool (ConvP) block as shown in Figure~\ref{fig:fused}. FC blocks each consist of a fully connected layer with $m$ nodes for some $m$, batch normalization and binary activation. ConvP blocks each consist of a convolutional layer with $f$ filters for some $f$, a pooling layer and batch normalization and binary activation. A convolution layer has a kernel of size 3x3 with stride 1 and padding 1. A pooling layer has a kernel of size 3x3 with stride 2 and padding 1.

\begin{figure}
    \centering
    \includegraphics[width=\linewidth]{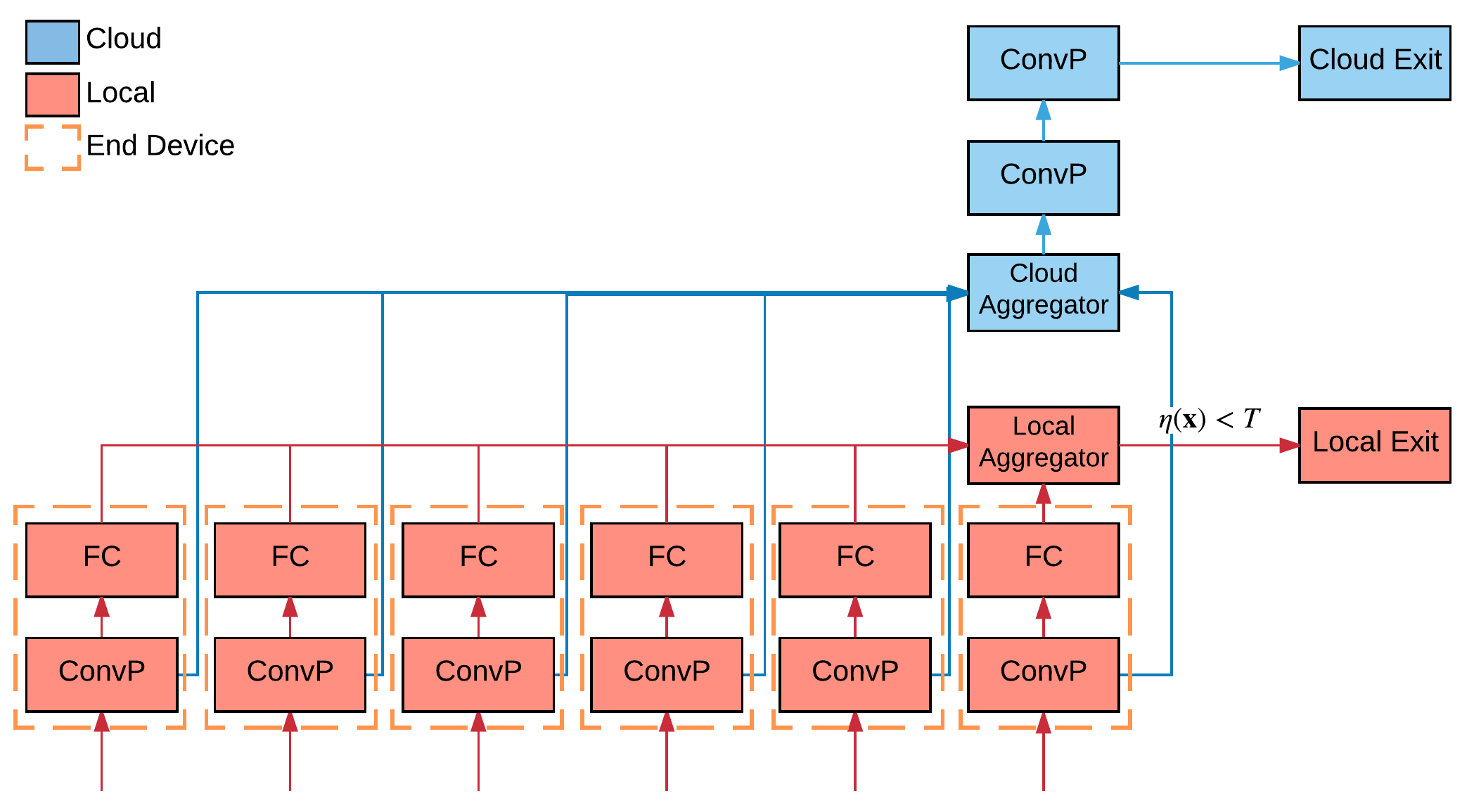}
    \caption{The DDNN architecture used in the system evaluation. The FC and ConvP blocks in red and blue correspond to layers run on end devices and the cloud respectively. The dashed orange boxes represent the end devices and show which blocks of the DDNN are mapped onto each device. The local aggregator shown in red combines the exit output (a short vector with length equal to the number of classes) from each end device in order to determine if local classification for the given input sample can be performed accurately. If the local exit is not confident (\ie~$\eta(x) > T$), the activation output after the last convolutional layer from each end device is sent to the cloud aggregator (shown in blue), which aggregates the input from each device, performs further NN layer processing, and outputs a final classification result. The aggregation of input for multiple end devices is discussed in Section~\ref{sec:agganalysis}.}
    \label{fig:merge}
\end{figure}

For our experiments, we use version (c) from Figure~\ref{fig:ddnn}, with six end devices. The system presented can be generalized to a more elaborated structure which includes an edge layer, as shown in (d), (e) or (f) of Figure~\ref{fig:ddnn}. Figure~\ref{fig:merge}~depicts a detailed view of the DDNN system used in our experiments. In this system, we have six end devices shown in red, a local aggregator, and a cloud aggregator. During training, output from each device is aggregated together at each exit point using one of the aggregation schemes described in Section~\ref{sec:agg}. We provide detailed analysis on the impact of aggregation schemes at both the local and cloud exit points in Section~\ref{sec:agganalysis}. All DDNNs in our experiments are trained with Adam~\cite{kingma2014adam} using the following hyper-parameter settings: $\alpha$ of 0.001, $\beta_1$ of 0.9, $\beta_2$ of 0.999, and $\epsilon$ of 1e-8. We train each DDNN for 100 epochs. When training the DDNN, we use equal weights for the local and cloud exit points. We explored heavily weighting both the local exit and the cloud exit, but neither weighting scheme significantly changed the accuracy of the system. This indicates that this solution to the dataset and the problem we are exploring is not sensitive to the weights, but this may not be true for other datasets and problems\footnote{In GoogleNet~\cite{szegedy2015going}, a less than 1\% difference in accuracy was observed based on the values of the weight parameters}.

\subsection{Multi-view Multi-camera Dataset}
\label{sec:dataset}
We evaluate the proposed DDNN framework on a multi-view multi-camera dataset~\cite{roig2011conditional}. This dataset consists of images acquired at the same time from six cameras placed at different locations facing the same general area. For the purpose of our evaluation, we assume that each camera is attached to an end device, which can transmit the captured images over a bandwidth-constraint wireless network to a physical endpoint connected to the cloud. 

\begin{figure}
    \centering
    \includegraphics[width=\linewidth]{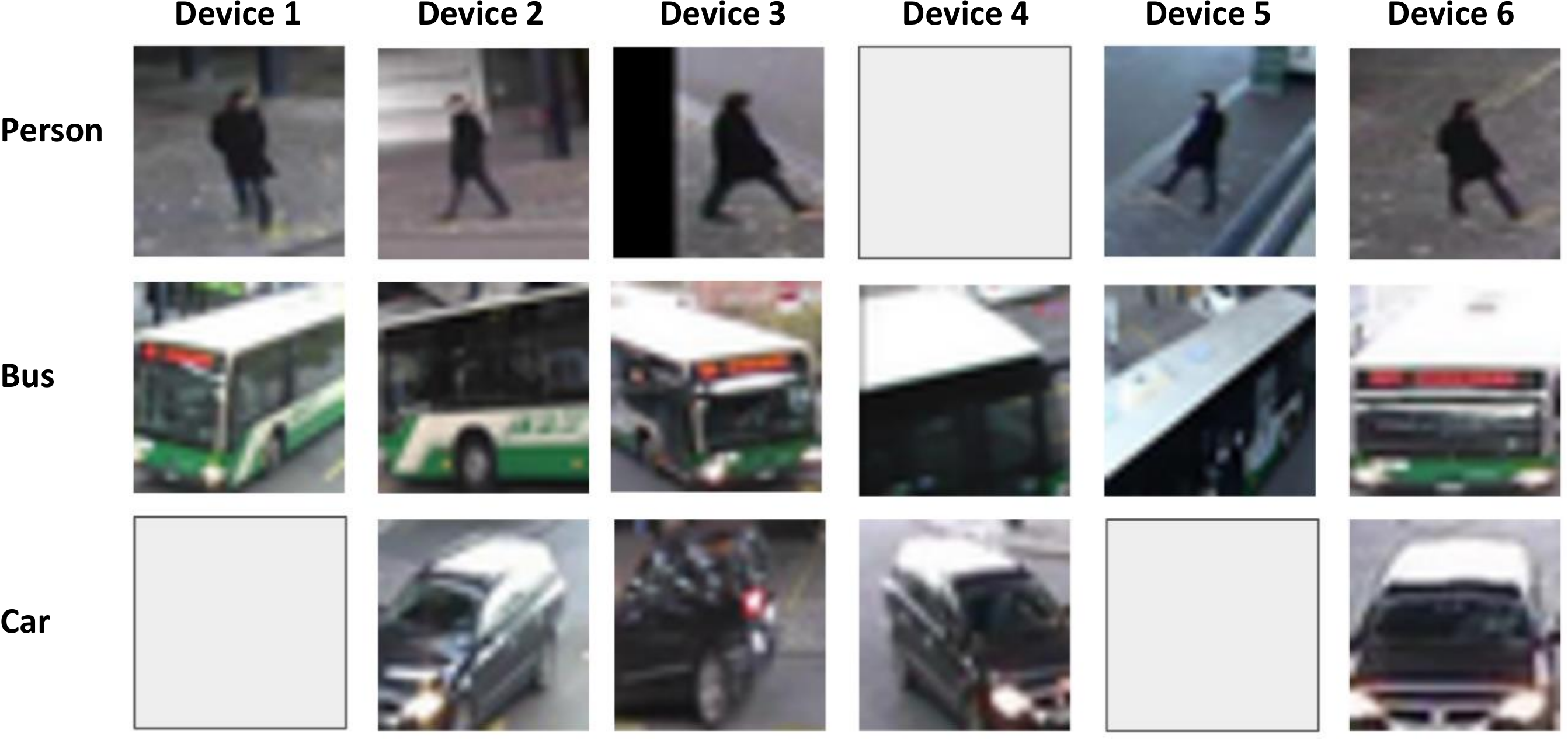}
    \caption{Example images of three objects (person, bus, car) from the multi-view multi-camera dataset. The six devices (each with their own camera) capture the same object from different orientations. An all grey image denotes that the object is not present in the frame.}
    \label{fig:multiview}
\end{figure}

\begin{figure}
    \centering
    \includegraphics[width=\linewidth]{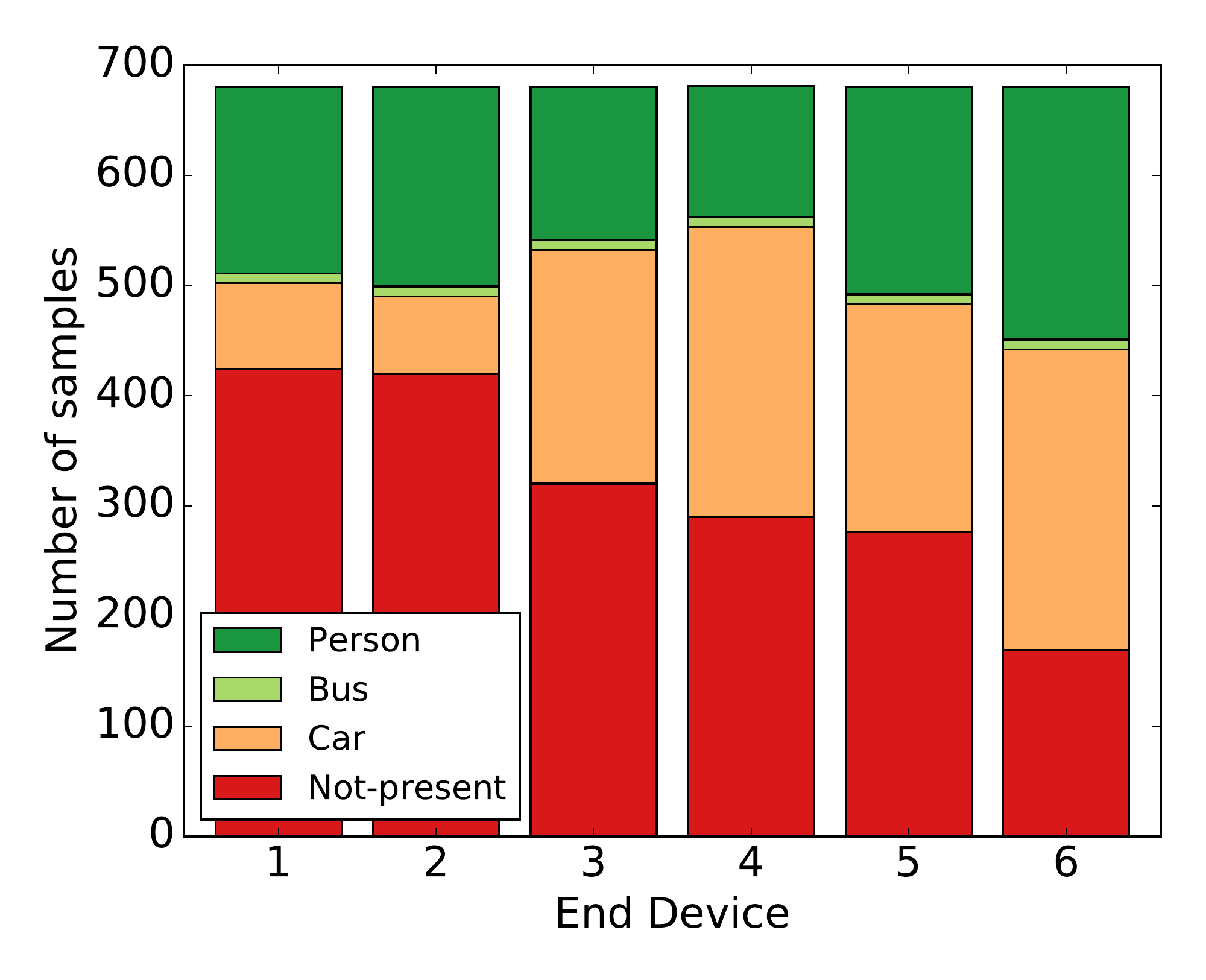}
    \caption{The distribution of class samples for each end device in the multi-view multi-camera dataset.}
    \label{fig:dataset}
\end{figure}

The dataset provides object bounding box annotations. Multiple bounding boxes may exist in a single image, each of which corresponds to a different object in the frame. In preparing the dataset, for each bounding box, we extract an image, and manually synchronize\footnote{In practical object tracking systems, this synchronization step is typically automated~\cite{li2013survey}.} the same object across the multiple devices that the object appears in for the given frame. Examples of the extracted images are shown in Figure~\ref{fig:multiview}. Each row corresponds to a single sample used for classification. We resize each extracted sample to a 32x32 RGB pixel image. For each device that a given object does not appear in, we use a blank image and assign a label of -1, meaning that the object is not present in the frame. Labels 0, 1, and 2 correspond to car, bus and person, respectively. Objects that are not present in a frame (\ie,~label of -1) are not used during training. We split the dataset into $680$ training samples and $171$ testing samples. Figure~\ref{fig:dataset} shows the distribution of samples at each device. Due to the imbalanced number of class samples in the dataset, the individual accuracy of each end device differs widely, as shown by the ``Individual" curve of Figure~\ref{fig:increasing}. A full description of the training process for the individual NN models is provided in Section~\ref{sec:scale-device}. The processed dataset used in this paper is available at~\cite{mvmc}.

\subsection{Impact of Aggregation Schemes}
\label{sec:agganalysis}
In order to perform classification on the input from multiple end devices, we must aggregate the information from each end device. We consider three aggregation methods (max pooling, average pooling, and concatenation) outlined in Section~\ref{sec:agg}, at both the local and cloud exit points. The accuracy of different aggregation schemes are shown in Table~\ref{tab:aggregate_schemes}. The first two letters identify the local aggregation scheme and the last two letters identify the scheme used by the cloud aggregator. For example, MP-CC means the local aggregator uses max-pooling and the cloud uses concatenation. Recall that each input to the local aggregator is a floating-point vector of length equal to the number of classes (corresponding to the output from the final FC block for a single device as shown in Figure~\ref{fig:merge}) and the device output sent to the cloud aggregator is the output from the final ConvP block.

\begin{table}[t]
\centering
\caption{Accuracy of aggregation schemes. The first two letters identify the local aggregation scheme, and the last two letters identify the cloud aggregation scheme. For example, MP-CC means the local aggregator uses max-pooling and the cloud aggregator uses concatenation. The accuracy of each exit point (either local or cloud) is computed using the entire test set. In practice, we will exit a portion of samples locally based on the entropy threshold $T$ and send the remaining samples in the cloud. Due to its high performance, MP-CC is used in the remaining experiments of this paper.}
\label{tab:aggregate_schemes}
\begin{tabular}{|l|l|l|}
\hline
\textbf{Schemes} & \textbf{Local Acc. (\%)} & \textbf{Cloud Acc. (\%)}   \\
\hline
MP-MP   & 95           &	91                              \\
\textbf{MP-CC}  & \textbf{98}  &	\textbf{98}                     \\
AP-AP   & 86           &	98                              \\
AP-CC   & 75           &	96                              \\
CC-CC   & 85           &	94                              \\
AP-MP   & 88           &	93                              \\
MP-AP   & 89           &	97                              \\
CC-MP   & 77           &	87                              \\
CC-AP   & 80           &	94                              \\
\hline
\end{tabular}
\end{table}

The MP-MP scheme has good classification accuracy for the local aggregator but poor performance in the cloud. The elements in the vectors at the local aggregator correspond to the same features (\eg,~the first item is the likelihood that the input corresponds to that class). Therefore, max pooling corresponds to taking the max response for each class over all end devices, and shows good performance. On the other hand, since the information sent from the end devices to the cloud is the activation output from the filters at each device, which corresponds to different visual features in the input from the viewpoint of each individual end device, max pooling these features does not perform well. 

Comparing MP-MP and MP-CC schemes, though both use MP for local aggregators, MP-CC increases the accuracy of the local classifier. In the training phrase, during backpropagation the MP-MP scheme only passes gradients through a device that gives the highest response while MP-CC scheme passes gradients through all devices. Therefore, using CC aggregator in the cloud allows all devices to learn better filters (filter weights) that give a stronger response for the local MP aggregator, resulting in a better classification accuracy.

The CC-CC scheme shows an opposite trend where the local accuracy is poor while the cloud accuracy is high. Concatenating the local information (instead of a pooling scheme), does not enforce any relationship between output for the same class on multiple devices and therefore performs worse. Concatenating the output for the cloud aggregator maintains the most information for NN layer processing in the cloud and therefore performs well. 

Generally, for the local aggregator, average pooling performs worse than max pooling. This is because some of the end devices do not have the object present in the given frame. Average pooling take average of all outputs from end devices; this compromises the strong outputs from end devices in which the object is present. Based on these results, we use the MP-CC aggregation scheme throughout the paper.

\subsection{Entropy Threshold}
\label{sec:entropy}
\begin{figure}
    \centering
    \includegraphics[width=\linewidth]{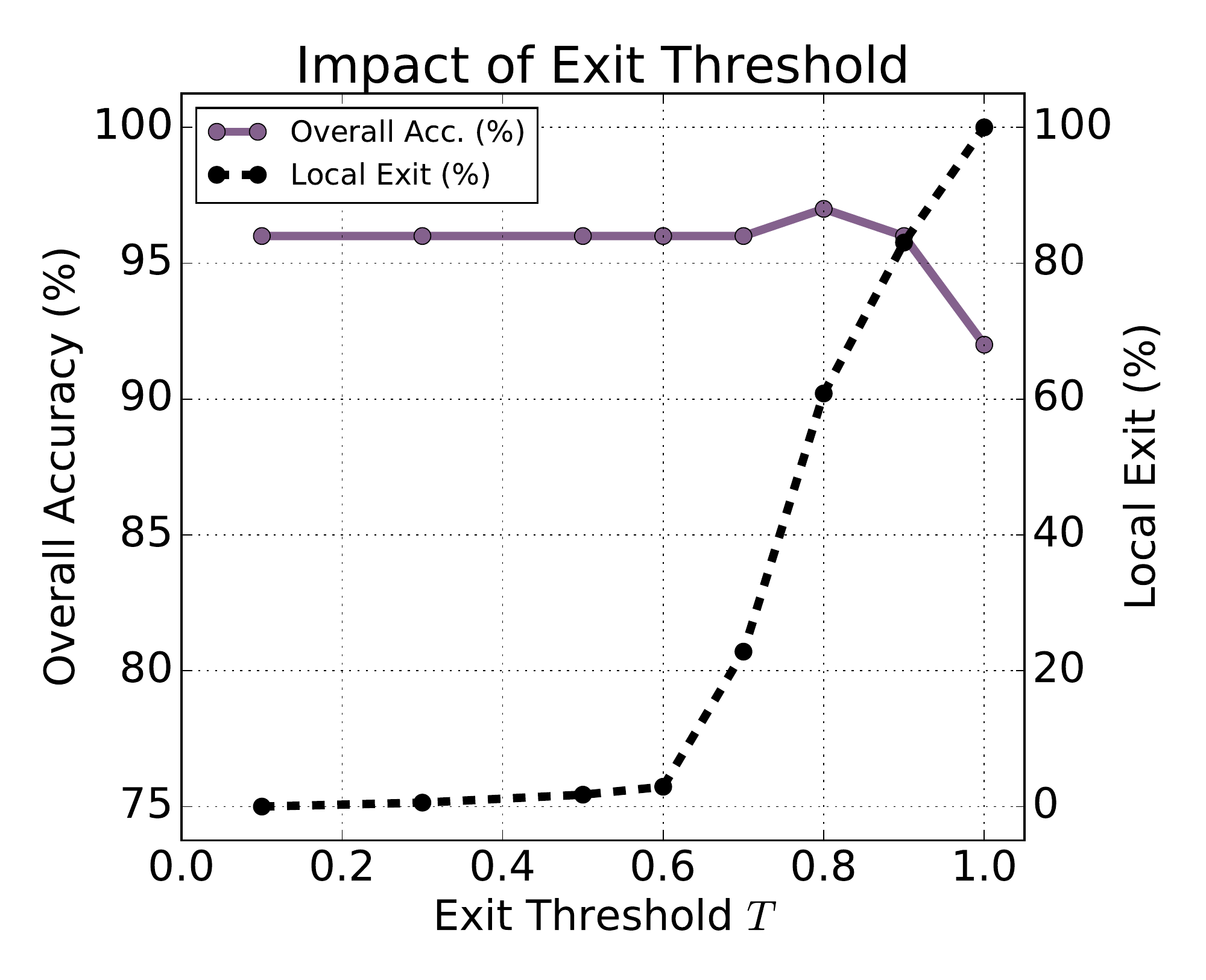}
    \caption{Overall accuracy of the system as the entropy threshold for the local exit is varied from 0 to 1. For this experiment, 4 filters are used in the ConvP blocks on the end devices.}
    \label{fig:thresholdvsacc}
\end{figure}

The entropy threshold for an exit point, $T$, corresponds to the level of confidence that is required in order to exit a sample. A threshold value of $0$ would mean that no samples will exit and a value of $1$ would mean that all samples exit at that point.  Figure~\ref{fig:thresholdvsacc} shows the relationship between $T$ at the local aggregator and the overall accuracy of the DDNN. We observe that as more samples are exited at the local exit, the overall accuracy decreases. This is expected, as the accuracy of the local exit is typically lower than that of the cloud exit. 

We need to set the threshold appropriately to achieve a balance between the communication cost, as defined in Section~\ref{sec:ddnn-comm}, latency and accuracy of the system. In this case, we see that setting the threshold to $0.8$ results in the best overall accuracy with significantly reduced communication, \ie,~97\% accuracy while exiting 60.82\% of samples locally as shown in Table~\ref{tab:threshold} where in addition to local exit (\%) and overall accuracy (\%), communication cost in bytes is given. We set $T=0.8$ for the remaining experiments in the system evaluation, unless noted otherwise. 


The local classifier may do better than cloud for certain samples where low-level features are more robust in classification than higher-level features. By setting an appropriate threshold $T$, we can improve overall accuracy. In this experiment, $T=0.8$ corresponds to that sweet spot where some samples which are incorrectly classified by the cloud classifier can actually be correctly classified by the local classifier. Such a threshold indicates the optimal point where both local and cloud classifier work best together.



\begin{table}[H]
\centering
\caption{Effects of different exit threshold ($T$) settings for the local exit. $T=0.8$ is used in the remaining experiments.}
\label{tab:threshold}
\begin{tabular}{|l|l|l|l|l|l|}
\hline
\textbf{$T$} & \textbf{Local Exit (\%)} & \textbf{Overall Acc. (\%)} & \textbf{Comm. (B)}\\
\hline
    0.1   & 0.00    &  96  &                            140           \\
    0.3   & 0.58    &  96  &                            139           \\
    0.5   & 1.75    &  96  &                            138           \\
    0.6   & 2.92    &  96  &                            136           \\
    0.7   & 22.81   &  96  &                            111           \\
    \textbf{0.8}   & \textbf{60.82}   &  \textbf{97}  & \textbf{62}  \\
    0.9   & 83.04   &  96  &                            34           \\
    1.0   & 100.00  &  92  &                            12          \\
\hline
\end{tabular}
\end{table}

\subsection{Impact of Scaling Across End Devices}
\label{sec:scale-device}
\begin{figure}
    \centering
    \includegraphics[width=\linewidth]{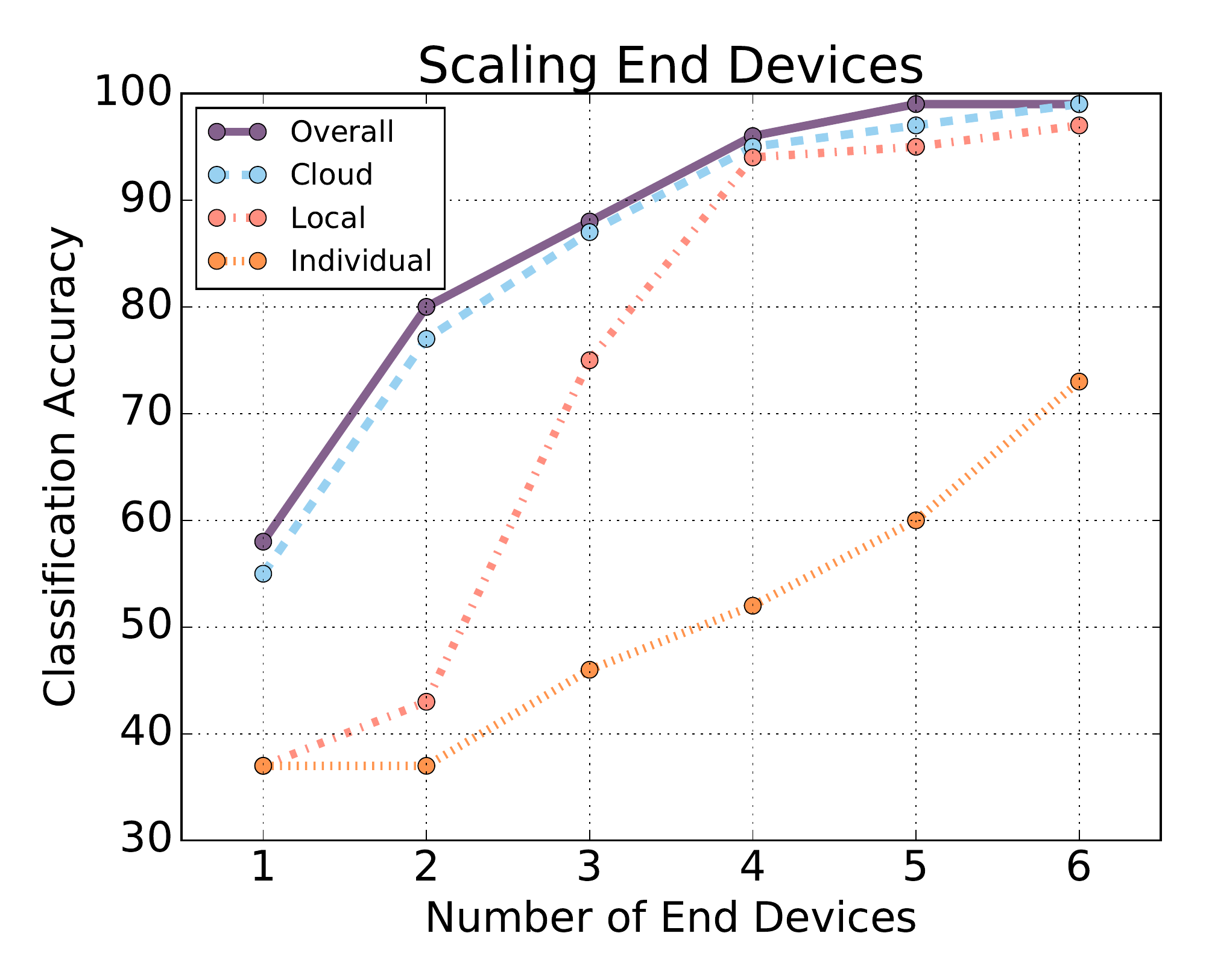}
    \caption{Accuracy of the DDNN system as additional end devices are added. The accuracy of ``Overall'' is obtained by exiting a percentage of the samples locally and the rest in the cloud. The accuracy of ``Cloud'' and ``Local'' are computed by exiting all samples at each point, respectively. The end devices are ordered by their ``Individual'' classification accuracy, sorted from worst to best.}
    \label{fig:increasing}
\end{figure}

In order to scale DDNNs across multiple end devices, we distribute the lower sections of Figure~\ref{fig:merge}, shown in red, over the corresponding devices, outlined in orange. Figure~\ref{fig:increasing} shows how the accuracy of the system improves as additional end devices (each with its attached input cameras) are added. The devices are added in order sorted by their individual accuracy from worst to best (\ie,~the device with the lowest accuracy first and the device with the highest accuracy last).

The first observation is the large variation in the individual accuracy of the end devices, as noted earlier. Due to the nature of the dataset, some devices are naturally better positioned and generally have clearer observations of the objects. Looking at the viewpoints of each camera in Figure~\ref{fig:multiview}, we see that the selected examples for Device 6 have clear frontal views of each object. This viewpoint gives Device 6 the highest individual accuracy at over 70\%. By comparison, Device 2 has the lowest individual accuracy at under 40\%. 

The ``Local'' and ``Cloud'' curves show the accuracy of the system at each exit point when all samples are exited at that point. We observe that the cloud exit point outperforms the local exit point at all numbers of end devices. The gap is widest when there are fewer devices. This suggests that the additional NN layers in the cloud significantly improve the final classification result when the problem is more difficult due to limited labeled training data for an end device. Once all six end devices are added, both the local and cloud aggregators have high accuracy. The ``Overall'' curve represents the overall accuracy of the system when the threshold for the local exit point is set to $0.8$. We see that this curve is roughly equivalent to exiting all samples at the cloud (but at a much reduced communication cost as 60.82\% of samples are exited locally). Generally, these results show that by combining multiple viewpoints we can increase the classification accuracy at both the local and cloud level by a substantial margin when compared to the individual accuracy of any device. The resulting accuracy of the DDNN system is superior to any individual device accuracy by over 20\%. Moreover, we note that the 60.82\% of samples which exit locally enjoy lowered latency in response time.

\subsection{Impact of Cloud Offloading on Accuracy Improvements}
\begin{figure}
    \centering
    \includegraphics[width=\linewidth]{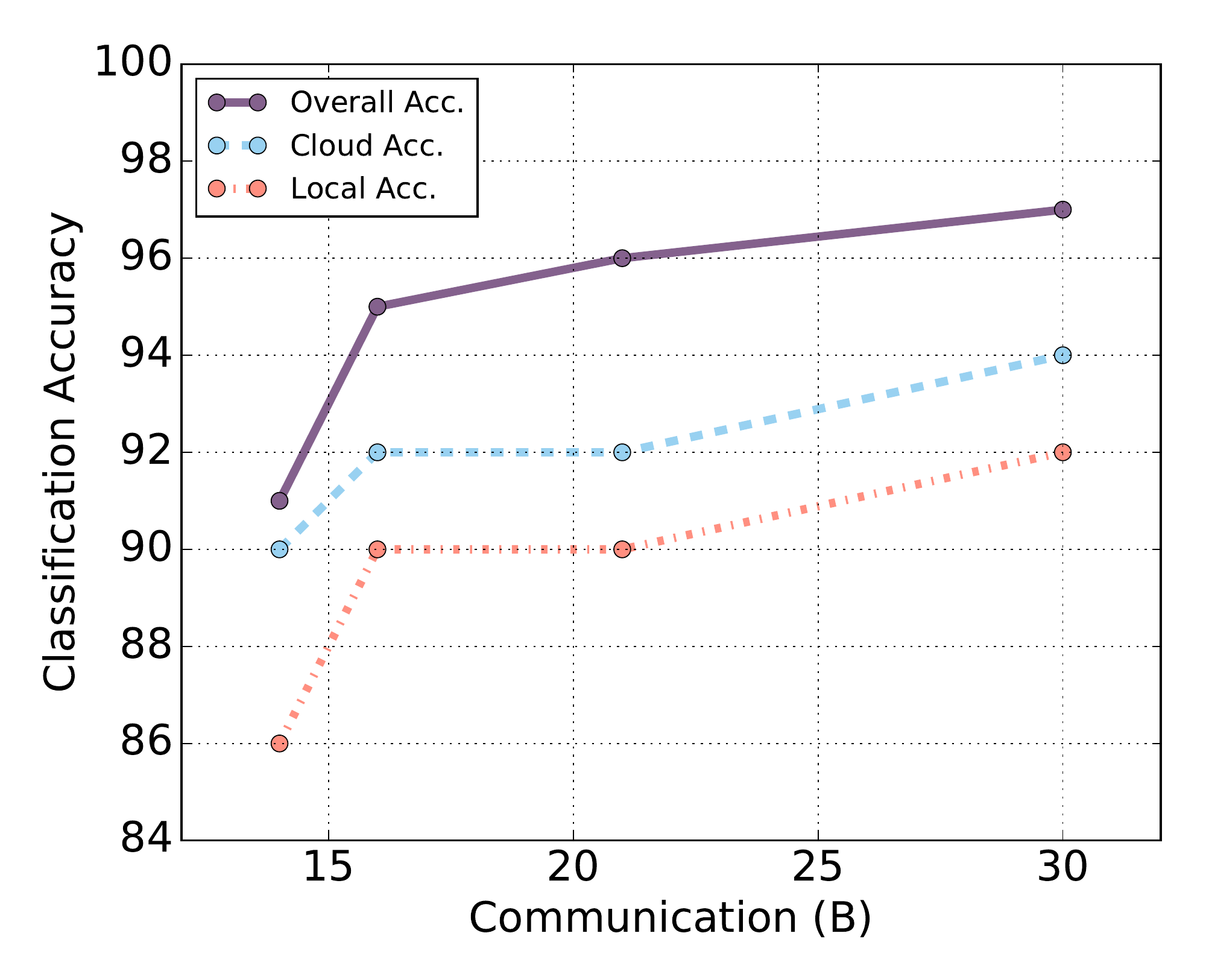}
    \caption{Accuracy and communication cost (in bytes) for increasingly larger end device memory sizes that accommodate additional filters. We notice that cloud offloading leads to improved accuracy.}
    \label{fig:commvsacc}
\end{figure}

DDNNs improve the overall accuracy of the system by offloading difficult samples to the cloud, which perform further NN layer processing and final classification.  Figure~\ref{fig:commvsacc} shows the accuracy and communication costs of DDNN as the number of filters on the end devices increases. For all settings, the NN layers stored on an end device require under $2$ KB of memory. In this experiment, we configure the local exit threshold $T$ such that around $75\%$ of samples are exited locally and around $25\%$ of samples are offloaded to the cloud. We see that DDNNs achieve around a $5\%$ improvement in accuracy compared to using just the local aggregator. This demonstrates the advantage for offloading to the cloud even when larger models (more filters) with improved local accuracy are used on the end devices.

\subsection{Fault Tolerance of DDNNs}
\begin{figure}
    \centering
    \includegraphics[width=\linewidth]{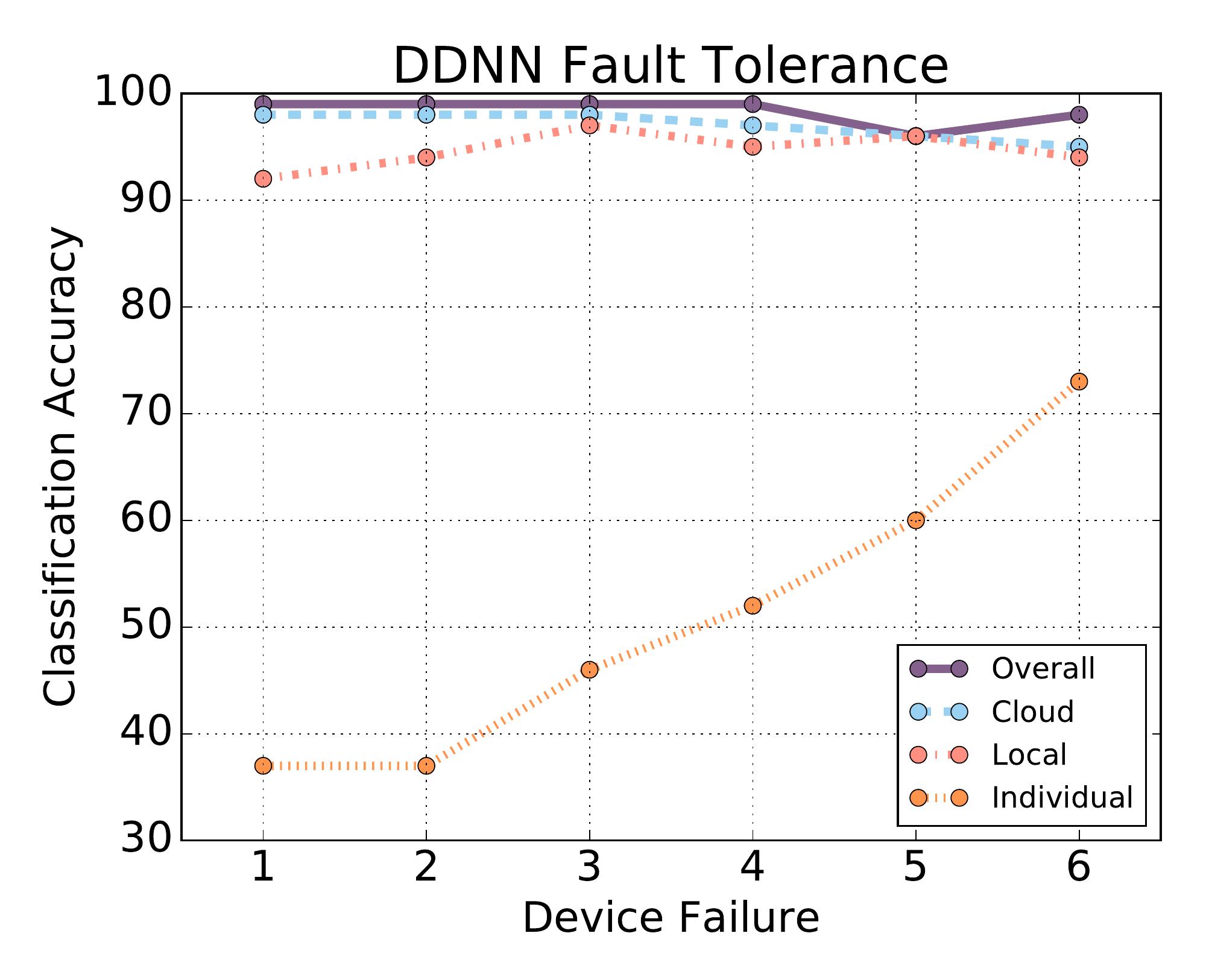}
    \caption{The impact on DDNN system accuracy when any single end device has failed.}
    \label{fig:fault}
\end{figure}

A key motivation for distributed systems is fault tolerance. Fault tolerance implies that the system still works well even when some parts are broken. In order to test the fault tolerance of DDNN, we simulate end device failures and look at the resulting accuracy of the system. Figure~\ref{fig:fault} shows the accuracy of the system under the presence of individual device failures. Regardless of the device that is missing, the system still achieves over a $95\%$ overall classification accuracy. Specifically, even when the device with the highest individual accuracy has failed, which is Device 6, the overall accuracy is reduced by only $3\%$. This suggests that for this dataset, the automatic fault tolerance provided by DDNN makes the system reliable even in the presence of device failure.

We can also view figure~\ref{fig:increasing} from the perspective of providing fault tolerance for the system. As we decrease the number of end devices from 6 to 4, we observe that the overall accuracy of the system drops only $4\%$. This suggests that the system can also be robust to mutliple failing end devices.

\subsection{Reducing Communication Costs}
\label{sec:comm}
DDNNs significantly reduces the communication cost of inference compared to the standard method of offloading raw sensor input to the cloud. Sending a 32x32 RGB pixel image (the input size of our dataset) to the cloud costs 3072 bytes per image sample. By comparison, as shown in Table~\ref{tab:threshold}, the largest DDNN model used in our evaluation section requires only 140 bytes of communication per sample on average (an over 20x reduction in communication costs). This communication reduction for an end device results from transmitting class-label related intermediate results to the local aggregator for all samples and binarized communication with the cloud when additional NN layer processing is required for classification with improved accuracy.

\section{DDNN Provision for\\Horizontal and Vertical Scaling}
The evaluation in the previous section shows that DDNN is able to achieve high overall accuracy through provisioning the network to scale both horizontally, across end devices, and vertically, over the network hierarchy. Specifically, we show that DDNN scales vertically, by exiting easier input samples locally for low-latency response and offloading difficult samples to the cloud for high overall recognition accuracy, while maintaining a small memory footprint on the end devices and incurring a low communication cost. This result is not obvious, as we need sufficiently good feature representations from the lower parts of the DNN (running on the end devices with limited resources) in order for the upper parts of the neural network (running in the cloud) to achieve high accuracy under the low communication cost constraint. Therefore, we show in a positive way that the proposed method of jointly training a single DNN with multiple exit points at each part of the distributed hierarchy allows us to meet this goal. That is, DDNN optimizes the lower parts of the DNN to create a sufficiently good feature representations to support both samples exited locally and those processed further in the cloud.

To meet the goal of horizontal scaling, we provide a principled way of jointly training a DNN with inputs from multiple devices through feature pooling via local and cloud aggregators and demonstrate that by aggregating features from each device we can dramatically improve the accuracy of the system both at the local and cloud level. Filters on each device are automatically tuned to process the geographically unique inputs and work together toward to the same overall objective leading to high overall accuracy. Additionally, we show that DDNN provides built-in fault tolerance across the end devices and is still able to achieve high accuracy in the presence of failed devices. 

\section{Conclusion}
In this paper, we propose a novel distributed deep neural network architecture (DDNN) that is distributed across computing hierarchies, consisting of the cloud, the edge and end devices. We demonstrate for a multi-view, multi-camera dataset that DDNN scales vertically from a few NN layers on end devices or the edge to many NN layers in the cloud and scales horizontally across multiple end devices. The aggregation of information communicated from different devices is built into the joint training of DDNN and is handled automatically at inference time. This approach simplifies the implementation and deployment of distributed cloud offloading and automates sensor fusion and system fault tolerance.

The experimental results suggest that with our DDNN framework, a single DNN properly trained can be mapped onto a distributed computing hierarchy to meet the accuracy, communication and latency requirements of a target application while gaining inherent benefits associated with distributed computing such as fault tolerance and privacy. 

DDNNs reduce the required communication compared to a standard cloud offloading approach by exiting many samples at the local aggregator and sending a compact binary feature representation to the cloud when additional processing is required. For our evaluation dataset, the communication cost of DDNN is reduced by a factor of over 20x compared to offloading raw sensor input to a DNN in the cloud which performs all of the inference computation.

DDNN provides a framework for further research in mapping DNN into a distributed computing hierarchy. For future work, we will investigate the performance of DDNNs on applications with a larger dataset with multiple types of input modalities~\cite{cha2015multimodal} and more end devices. Currently, all layers in DDNN are binary. While binary layers are a requirement for end devices due to the limited space on devices, it is not necessary in the cloud. We will explore other types of aggregation schemes and mixed precisions schemes where the end devices use binary NN layers and the cloud uses mixed-precision or floating-point NN layers.

\section*{Acknowledgment}
This work is supported in part by gifts from the Intel Corporation
and in part by the Naval Supply Systems Command
award under the Naval Postgraduate School Agreements No.
N00244-15-0050 and No. N00244-16-1-0018.

\bibliographystyle{IEEEtran}
\bibliography{IEEEabrv,sigproc}
\end{document}